# ParaTaxe: a Suite of Tools for Processing and Publishing Multi-parallel Corpora

Olivier Kraif[1] , Cyrille François[2] , Elnaz Jalilian[1] , Natacha Rimasson-Fertin[3] , Pascale Roux[4]

[1] LIDILEM, Univ. Grenoble Alpes – olivier.kraif@univ-grenoble-alpes.fr, elnazjalilian@gmail.com

[2] CIEL, University of Lausanne – cyrille.francois@unil.ch

[3] ILCEA4, Univ. Grenoble Alpes –natacha.rimasson-fertin@univ-grenoble-alpes.fr

[4] Passages XX-XXI, Univ. Lyon 2 – pascale.roux@univ-lyon2.fr

## Abstract

Over the past fifteen years, digital tools for parallelizing and aligning textual corpora have developed significantly within the field of digital humanities. Several projects and software tools have been created to compare different translations or versions of texts. Despite these technical advances, many parallel corpora produced in academic contexts quickly become inaccessible online due to the lack of maintenance or funding. To address this gap, we present a suite of open-source tools developed within the ACR Grimm Tradalign and ParaTaxe projects. This suite includes the AIlign aligner, based on multilingual embeddings, a Python processing pipeline for converting and aligning corpora, a web interface called ParaTaxe for managing collections of parallel corpora, and a web interface for exploring aligned corpora built with TEI-Publisher.

## Abstract

Au cours des quinze dernières années, les outils numériques permettant de paralléliser et d'aligner des corpus textuels se sont largement développés dans les humanités numériques. Plusieurs projets et logiciels ont été créés pour comparer différentes traductions ou versions de textes. Malgré ces avancées techniques, de nombreux corpus parallèles produits dans un cadre académique deviennent rapidement inaccessibles en ligne, faute de maintenance ou de financements. Pour répondre à cette lacune, nous présentons une suite d'outils open source développés dans les projets ACR Grimm Tradalign et ParaTaxe. Cet ensemble comprend l'aligneur AIlign, basé sur des embeddings multilingues, une chaîne de traitement Python pour convertir et aligner des corpus, une interface web de gestion de corpus nommée ParaTaxe, et une interface d'exploration de corpus alignés basée sur TEI-Publisher. Une première application a permis de publier un corpus parallèle des contes des frères Grimm.

# 1. Introduction

Over the past fifteen years, digital tools for corpus parallelization have developed significantly in the field of Digital Humanities: examples include the HyperMachiavel software (Gedzelman & Zancarini, 2011) for a corpus of translations of Machiavelli's *The Prince*, the alignment tool from Marianne Reboul's *Odysseus* project, which studies translations of *The Odyssey* into French from the 16th to the 20th century (Reboul, 2022), the CollateX software by Dekker & Middel (2011), which allows multiple alignments to be observed at the lexical level, and the Variance platform (see https://variance.unil.ch), dedicated to comparing different versions of the same edition.

Despite the maturity of alignment techniques and the large number of projects carried out in this field, it is clear that many parallel corpora developed in an academic context are not (or no longer) available online due to a lack of maintenance or funding to develop *ad hoc* sites (such as Hyperprince). For example, the CARMEL parallel corpus (El Bèze et al., 2006), a vast corpus of travel stories in four languages, was only accessible for two years, following the withdrawal of the industrial partner involved at the time. On the Ortolang Platform (https://www.ortolang.fr/), there are several parallel corpora available for download (GIEC corpus, UN Security Council Resolutions 1946-2015 corpus), but none of them can be consulted directly in a simple manner.

To fill this gap, we present a suite of open source tools developed as part of the *Grimm Tradalign* (funded by the ACR group) and *ParaTaxe* (funded by IDEX UGA) projects. There are four tools: the *AIlign* aligner (Kraif, 2025), a robust, state-of-the-art aligner based on multilingual embeddings; a Python processing chain that converts parallel corpora in TXT or DOCX format into XML-TEI, then aligns them (in TSV, TXT, and XML); a web interface called *ParaTaxe Dashboard* for managing collections of parallel corpora and launching this pipeline; and finally, a web interface for exploring aligned multilingual corpora, developed using TEI-Publisher, an instance of which has already been used to deploy a parallel corpus of Grimm's fairy tales (https://grimm.unil.ch/).

# 2. Main concepts and existing tools

By multi-parallel texts, we mean a generalization of the concept of parallel texts, which originally applies to pairs combining source texts and their translation into a target language. The alignment of such a parallel corpus is generally performed by an alignment software or an aligner, which, as formalized by Isabelle (1992), relies on two steps : *segmentation* (at varying degrees of granularity, paragraphs, sentences, phrases, or words) and *matching* equivalent segments. We also use the definition of parallelism given by Langé & Gaussier (1995), who based this concept on two criteria related to the structure of equivalence relations: 1/ quasi-bijectivity: segments of the source text generally have a corresponding segment in the target text; 2/ quasi-monotonicity: the order of the target segments generally respects that of the source sentences.

Based on these formal properties, we can generalize the concept of a multi-parallel corpus to a triplet <T,f,A>, where *T* is a collection of texts (which may be in one, two, or several languages), *f* is a segmentation function (which may operate at the level of paragraphs, sentences, verses, etc.), and *A* is a set of pairings between equivalent segments, with the concept of equivalence extended beyond that of translational equivalence: it can concern successive rewritings of the same text in the case of textual genetics or variations between several editions, different translations from the same source text, or even a mixture of these phenomena (as is the case in the Grimm corpus described below). The notion of a multi-parallel corpus *ultimately* becomes a formal property based on the constraints of quasi-bijectivity and quasi-monotonicity linked to alignment A.

For the publication of multi-parallel corpora defined in this way, there are two types of tools:

- Automatic aligners, which aim to produce alignments generally at the sentence level. Following the generation of surface aligners such as Gale & Church (1991), BMA (Moore, 2002) and Hunalign (Varga et al. 2007), semantic aligners based on sentence embeddings (vector representations derived from neural models) have shown greater accuracy and robustness with regard to phenomena such as parallelism breaks or significant distance between languages with different alphabets. Among these, Thomson & Koehn (2020), with the Vecalign system, and then Liu and Zhu (2022), with BertAlign, have produced a new state of the art. At the lexical level, we can mention statistical aligners such as FastAlign (Dyer et al. 2013), and more recently, based on transformer architectures, SimAlign (Jalili Sabet et al. 2020) or awesome-align (Dou & Neubig 2021). It should be noted that, with a few exceptions, such as the multi-aligner JAM (Kraif, 2015), these tools most often operate on pairs of texts. The WebAlignToolkit[1] interface provides access to various online aligners and allows users to obtain a sentence-level multi-alignment for two or more texts. In the field of aligning versions of the same text from a genetic perspective, the CollateX software (Dekker & Middel 2011) allows users to view correspondence graphs at the character string level. Finally, with Mkalign, Zimina & Fleury (2015) present a comprehensive environment for the alignment, visualization, and statistical exploration of aligned texts – the tool can be used locally on a desktop computer, but is not designed for online use.

- The second family of tools concerns interfaces for publishing parallel corpora. It is clear that a large number of parallel editions are based on *ad hoc* sites rather than generic tools: we can cite the examples of *HyperPrince*[2], *Translatoscope*[3], and *Odysseus Polymetis*[4]. There are platforms dedicated to the publication of different corpora, such as the very rich Opus (Tiedmann 2012), the Variance platform (Mahrer & Zufferey, 2019), and the collaborative Ugarit (Yousef et al. 2022), which allows users to edit word-by-word alignments and share them. Also noteworthy is the ParCoLab project (Stosic et al. 2024), which is based on the open source NoSketch Engine, but aims to publish a specific parallel corpus.

[1] Available at : http://phraseotext.univ-grenoble-alpes.fr/webAlignToolkit/

[2] Translations *of* Machiavelli's *The Prince*: http://hyperprince.ens-lyon.fr

[3] Translations of Aristophanes' Ploutos: https://elan.gricad-pages.univ-grenoble-alpes.fr/translatoscope

[4] Translations of Homer's Odyssey: https://odysseuspolymetis.github.io/paralogos/

In this article, we propose ParaTaxe, a suite of generic, open-source tools integrated into a single platform that allows users to align and publish their own editions of a multi-parallel corpus without investing in new developments.

## 3. The ParaTAXE suite

The ParaTAXE (Parallel Text Alignment eXploration and Edition) suite aims to integrate the entire workflow, from importing source files to viewing texts in parallel, with an unlimited number of columns (except for reading comfort, depending on screen size).

### *3.1 The Dashboard*

The dashboard is the corpus management interface[5]. Corpora are organized into *collections* (e.g., *Homère*), each of which contains *documents* (e.g., *Iliade1*), each corresponding to a list of *files* (e.g., *ILIADE1_1530_Samxon_fr.xml*). Files can be in different formats: DOCX, TXT, XML-TEI.

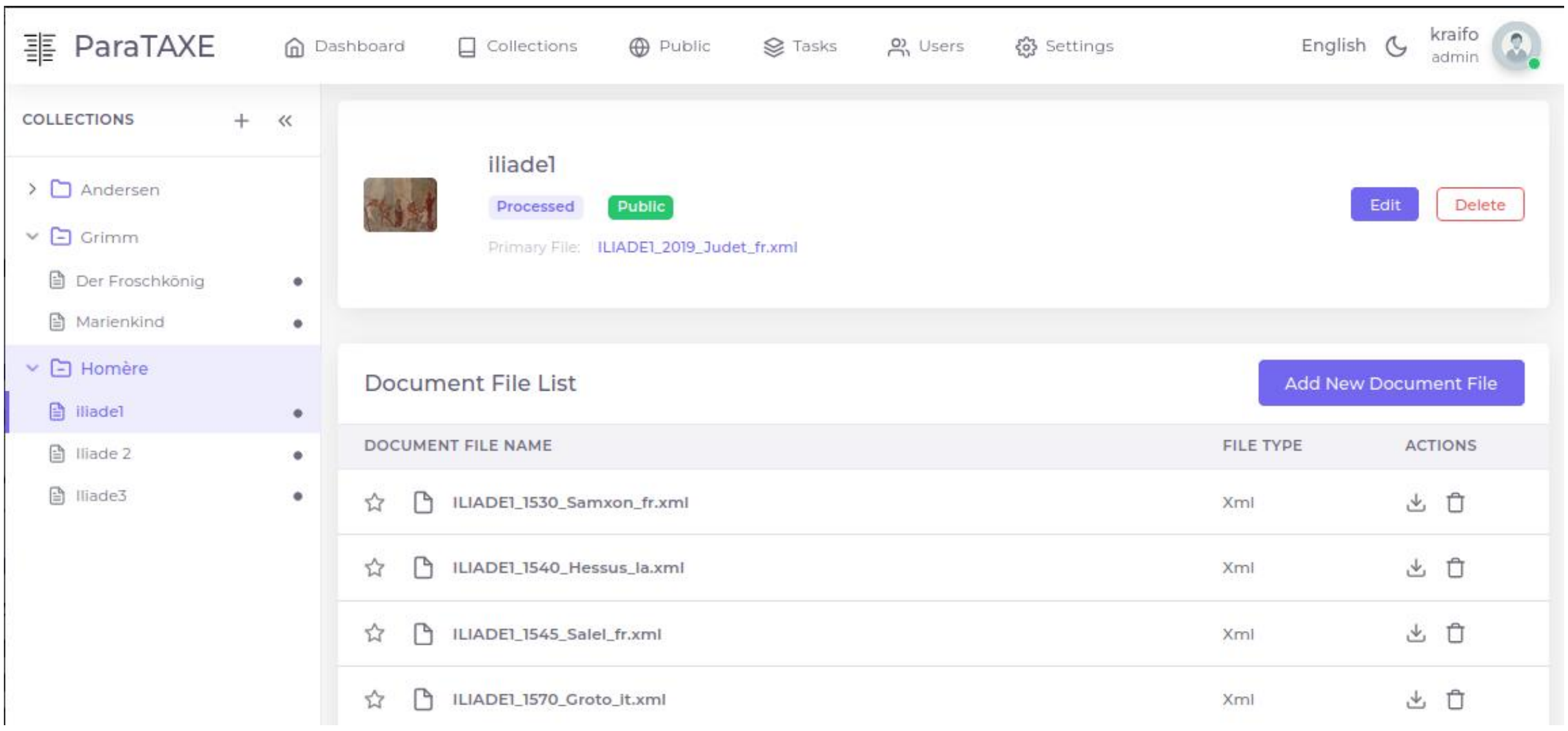

*Figure 1: Management of collections / documents / document files*

### *3.2 Launching the pipeline*

For each collection, a metadata file is defined to specify information such as ‘title’, ‘author’, ‘year’, ‘translator’, etc. for each file.

#### - File conversion

The pipeline involves conversion to XML-TEI if necessary, with metadata injection in the header and segmentation at the paragraph and sentence level. If the imported files are already in XML-TEI, it is necessary to indicate which segmentation tag should be taken into account for alignment (e.g., <l> for the verses of the Iliad).

[5] An instance of the tool is currently hosted here: http://phraseotext.univ-grenoble-alpes.fr:5000

**- Alignment**

Alignment is performed using the AIlign aligner (Kraif, 2025), a state-of-the-art aligner that performs a preliminary step of detecting alignable areas (e.g., by eliminating untranslated prefaces or chapters, or by identifying a different order in the publication of stories or short stories in a collection). The aligner itself is available on a git repository[6] for standalone use (command line interface). The aligned files are saved in different formats (TSV, TXT, XML-TEI) and can be downloaded by the user.

To simplify alignment, one file must be designated as the pivot: all other files are then aligned with the pivot (which implies, for *n* files, *n-1* aligned pairs).

### *3.3 Visualization*

Once the processing is complete[7], the documents can be viewed in a multi-parallel visualization. Clicking on a segment synchronizes the display with all other displayed versions (synchronization is achieved by alignment with the pivot, even if it does not appear in the columns).

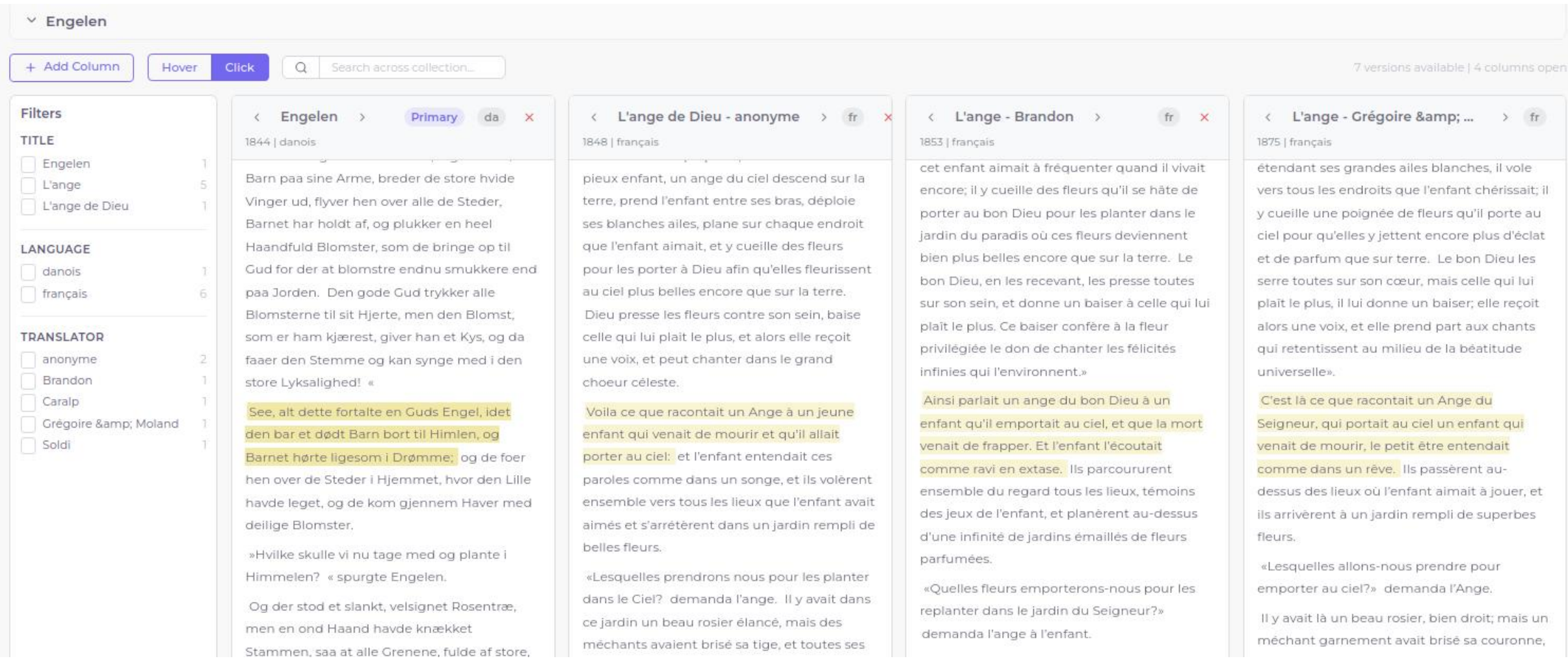

*Figure 2: Parallel display of aligned documents*

### *3.4 Publication of corpora*

Collections can be public or private: a public collection is accessible for consultation by other registered users of the platform. In addition, once the processing is complete, it is possible to make a document completely public, without authentication, by clicking on "Publish." This publication mode involves displaying an illustration icon, a presentation text, and multi-parallel columns. A search field allows the user to extract parallel concordances for a given expression.

[6] At the address: https://gricad-gitlab.univ-grenoble-alpes.fr/kraifo/ailign

[7] The pipeline jobs are executed using Redis, a queue management system for tasks in progress.

In addition, we have also developed TradAlign, a publishing interface based on TEI Publisher[8]. Independent of the ParaTAXE platform, it also allows multi-parallel publication and metadata search options. We used it to publish a translation corpus of the Grimm brothers' fairy tales[9] (Rimasson-Fertin et al. 2024). This corpus gather various editions of a selection of 10 Grimm tales, from the first edition of 1812 to the last of 1857, which has been taken as a pivot. French translations from the 19th century have been aligned with this pivot, as well as the previous original editions.

Note that the files obtained by the aligning pipeline of ParaTAXE are compatible with the format required for TradAlign.

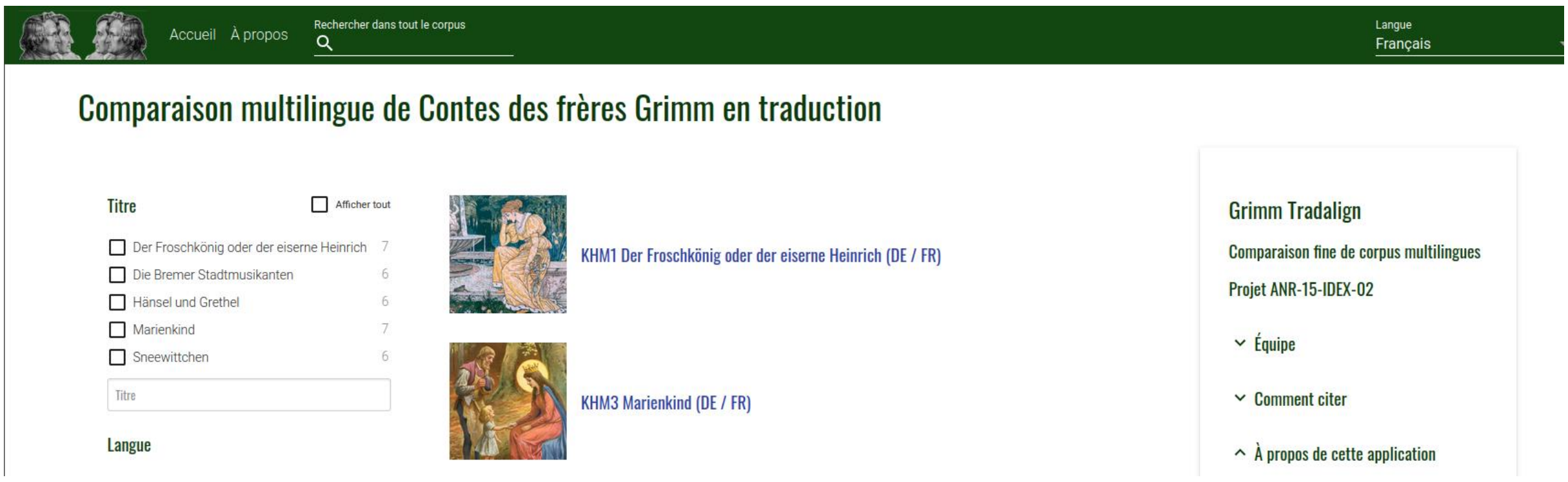

*Figure 3: Grimm TradAlign interface, based on TEI Publisher*

## 4. Conclusion and future works

The ParaTAXE suite is available on a Git repository under open source licence[10]. There are two branches: in the main branch, publication is automatically directed toward the TradAlign tool, which is integrated; in another lighter branch, without TEI-Publisher and Exist-db module, the aligned view is integrated into the dashboard.

The platform can be installed on a development or production server using a single command that creates docker containers[11].

In future versions, we plan to add a module (which is almost finished) for manual editing of automatic alignments, via a drag-and-drop system from a two-column view.

We also plan to add a lexical level alignment step (which already exists in AIlign) and a module to extract translation and textometric statistics (such as sentence lengths, vocabulary

[8] The implementation has been conducted by the Jinntec company under our supervision (https://jinntec.de/)

[9] The site is hosted by the University of Lausanne and can be viewed here: https://grimm.unil.ch

[10] Available here: https://gricad-gitlab.univ-grenoble-alpes.fr/jaliliel/parataxe-admin

[11] Docker is a technology that creates a kind of virtual environment to ensure the application's portability across different systems. The install procedure just consists in one command line (*sudo make dev* or *sudo make file*), with the *makefile* installing the dependencies and launching the various docker containers hosting the necessary services.

sizes, hierarchical indexes, occurrence and cooccurrence frequencies, correspondence frequencies, etc.)

## Acknowledgments

The Grimm corpus and the Grimm Tradalign interface were funded by ACR (Project ANR-15-IDEX-02) and UNIL. The ParaTAXE project received IDEX funding from UGA (IRGA2023 –ParaTAXE – G7H-IRG23D42).

## References

Dekker R. H., Middell G. (2011). Computer-Supported Collation with CollateX: Managing Textual Variance in an Environment with Varying Requirements. *Supporting Digital Humanities 2011*. University of Copenhagen, Denmark.

Dou Z.-Y., & Neubig G. (2021). Word alignment by fine-tuning embeddings on parallel corpora. In *Proceedings of the 16th Conference of the European Chapter of the Association for Computational Linguistics (EACL),* pp. 2112–2128.

Dyer C., Chahuneau V., Smith N. A. (2013). A simple, fast, and effective reparameterization of ibm model 2. In *Proceedings of the 2013 Conference of the North American Chapter of the Association for Computational Linguistics: Human Language Technologies*, pp. 644–648.

El-Bèze M., Richard C., Meyer R. (2006). CARMEL Project: Travel Stories. In *Proceedings of TALN 2006,* Louvain-la-Neuve.

Gedzelman S., Zancarini J.-C. (2011). HyperMachiavel: a translation comparison tool. *Lingua e stile*, 2011, vol. XLVI, no. 2, pp. 247-266.

Isabelle P. (1992). Bi-Textual Aids for Translators. In *Proceedings of the Eight Annual Conference of the UW Centre for the New OED and Text Research*, University of Waterloo, Waterloo, pp. 76-89,

Langé J.-M., Gaussier É. (1995). Alignement de corpus multilingues au niveau des phrases. *T.A.L.*, vol. 36, n. 1-2, pp. 67-79.

Jalili Sabet, M., Dufter, P., Yvon, F., & Schütze, H. (2020). SimAlign: High quality word alignments without parallel training data using static and contextualized embeddings. In *Findings of the Association for Computational Linguistics: EMNLP 2020* (pp. 1627–1643).

Kraif, O. (2015). Multialignement vs bialignement : à plusieurs, c'est mieux ! In *Actes de la 22e conférence sur le Traitement Automatique des Langues Naturelles (TALN 2015),* pp. 345–356.

Kraif O. (2025). Adaptive bi-textual alignment based on multilingual embeddings. *Proceedings of CORIA-TALN-RJCRI-RECITAL 2025*, Marseille, July 1-5, 2025. https://aclanthology.org/2025.jeptalnrecital-taln.4.pdf

Kraif O., Roux P. (2022). Comparison of an original text and its back-translations: what do textometric measures tell us?, in Meng Ji Christine and Michael Oakes (Eds.), *New methodologies in corpus-based translation studies: The empirical revolution in translation studies, Meta*, v. 67, no. 1, Les Presses de l'Université de Montréal: Montreal.

Liu L. & Zhu M. (2022). Bertalign: Improved word embedding-based sentence alignment for Chinese–English parallel corpora of literary texts. *Digital Scholarship in the Humanities*, Volume 38, Issue 2, June 2023, 621–634.

Mahrer R. & Zufferey J. (2019). *Variance: A digital platform for the study of textual rewriting across editions.* In W. Scholger & M. Büchler (Eds.), *Digital Scholarly Editions as Interfaces*. Norderstedt: Books on Demand.

Moore R. C. (2002). Fast and accurate sentence alignment of bilingual corpora. In *Conference of the Association for Machine Translation in the Americas,* pages 135–144, Tiburon, USA, October. Springer.

Stosic, D., Marjanović, S., Bernhard, D., Bach, X., Bras, M., Kevers, L., Retali-Medori, S., Vergez-Couret, M., & Werner, C. (2024). The ParCoLab parallel corpus and its extension to four regional languages of France. In N. Calzolari, M.-Y. Kan, V. Hoste, A. Lenci, S. Sakti, & N. Xue (Eds.), *Proceedings of the 2024 Joint International Conference on Computational Linguistics, Language Resources and Evaluation (LREC-COLING 2024)* (pp. 16014–16023).

Reboul M. (2022). Semi-automatic comparison of French translations of Homer's Odyssey (1547-1955). In Didier Alexandre and Glenn Roe (eds.), *Cultures et pratiques savantes du numérique*, Classiques Garnier ⟨10.48611/isbn.978-2-406-12961-5⟩

Rimasson-Fertin N., François C., Kraif O., Roux P., Jalilian E. (2024). *Grimm Tradalign: plateforme de comparaison de traductions de contes des frères Grimm*. UGA-UNIL.

Thomson B. & Koehn P. (2020). Vecalign: Improved Sentence Alignment in Linear Time and Space. P*roceedings of the 2019 Conference on Empirical Methods in Natural Language Processing and the 9th International Joint Conference on Natural Language Processing*, pages 1342–1348, Hong Kong, China, November 3–7, 2019.

Tiedemann J. (2012). Parallel Data, Tools and Interfaces in OPUS. In *Proceedings of the Eighth International Conference on Language Resources and Evaluation (LREC'12)*, pages 2214–2218, Istanbul, Turkey.

Varga D., Halácsy P., Kornai A., Nagy V., Németh L. & Trón V. (2007). Parallel corpora for medium density languages. *Amsterdam Studies In The Theory And History Of Linguistic Science Series* 4, 292:247.

Yousef T., Palladino C., Shamsian F., & Foradi M. (2022). Translation Alignment with Ugarit. *Information* (2022) 13(2): 65.

Zimina, M. & Fleury, S. (2015). Perspectives de l'architecture Trame/Cadre pour les alignements multilingues. *Nouvelles perspectives en sciences sociales*, 11(1), 325–353.